\documentclass[runningheads]{llncs}
\usepackage[T1]{fontenc}
\usepackage{graphicx}
\usepackage{subcaption}   
\usepackage{caption}      
\usepackage{multirow}
\usepackage{makecell}
\usepackage{float}
\usepackage{amsmath}
\usepackage{amssymb}
\usepackage{hyperref}
\usepackage{color}

\begin{document}

\title{Fusion2Print: Deep Flash--Non-Flash Fusion for Contactless Fingerprint Matching}
\titlerunning{Fusion2Print}
%

\author{Roja Sahoo
\and
Anoop Namboodiri
}
%
%
\institute{IIIT-Hyderabad, Gachibowli, India \\
\email{roja.sahoo@research.iiit.ac.in and anoop@iiit.ac.in}}

\maketitle
\vspace{-1em} 
\begin{figure}[!th]
    \centering
    \begin{subfigure}{0.55\textwidth}
        \centering
        \includegraphics[width=0.99\textwidth]{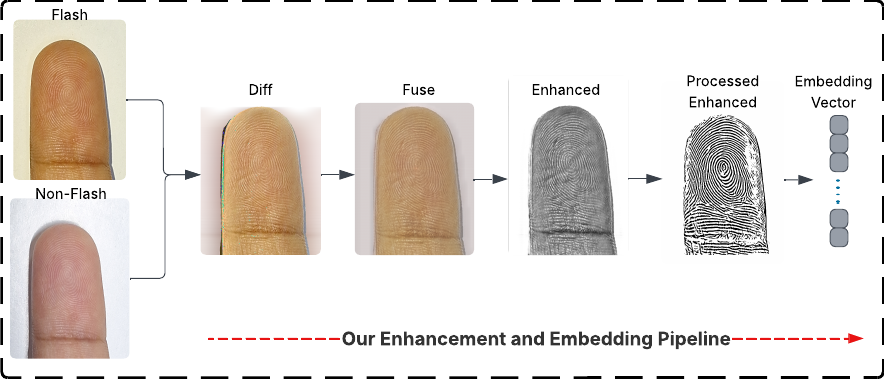}
        \caption{Flash--non-flash enhancement-embedding pipeline (Fusion2Print)}
        \label{fig:1a}
    \end{subfigure}
    \begin{subfigure}{0.44\textwidth}
        \centering
        \includegraphics[width=0.99\textwidth]{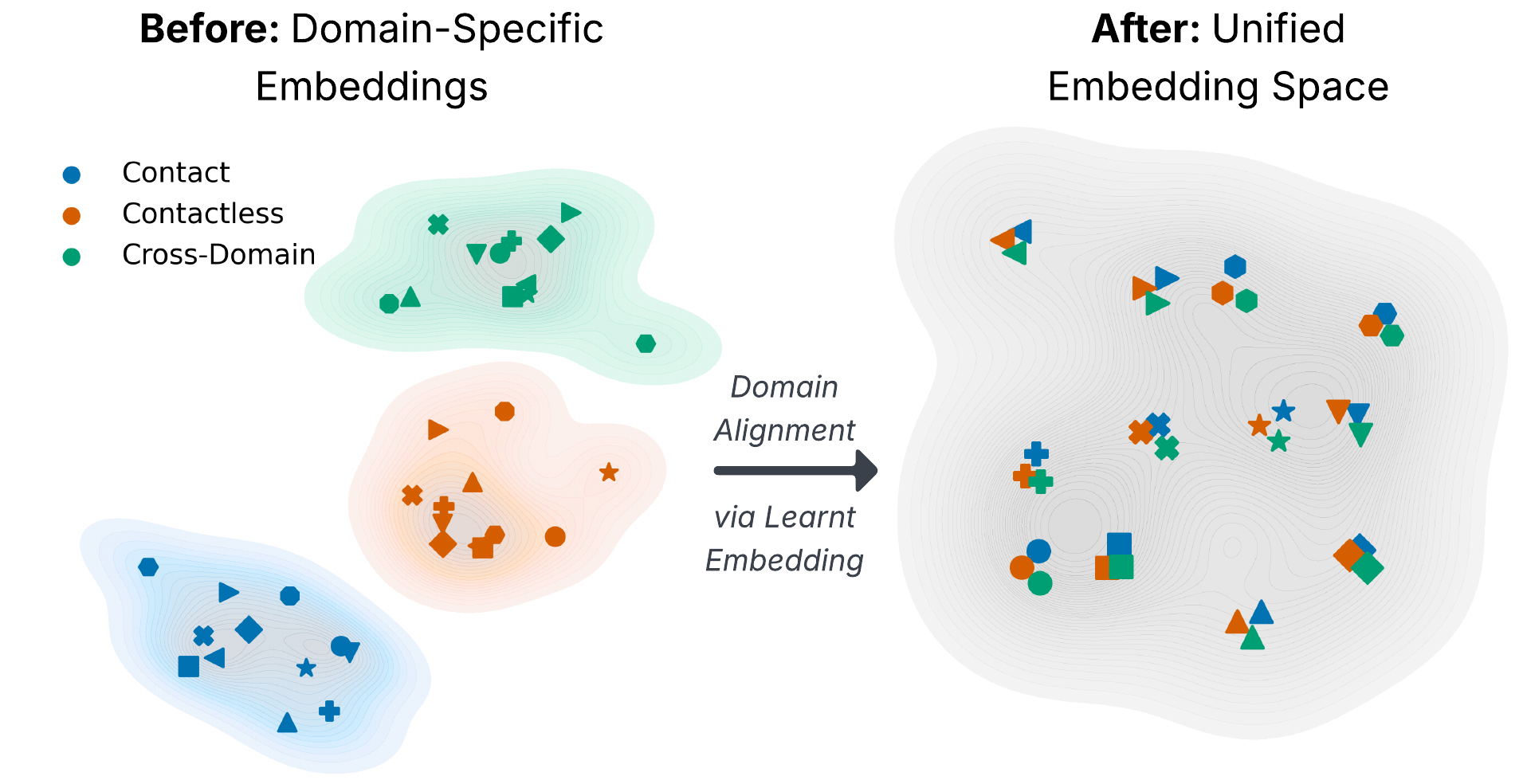}
        \caption{Embedding space before and after domain alignment}
        \label{fig:1b}
    \end{subfigure}

\caption{Overview of proposed fingerprint acquisition and enhancement framework: 
(a) The pipeline progressively enhances ridge-valley structure to produce an identity-discriminative embedding vector; (b) Embedding fine-tuning aligns domain-specific representations into a unified embedding space for contact and contactless fingerprints.}

\label{fig:1}
\end{figure}
\vspace{-3em}

\begin{abstract}
Contactless fingerprint acquisition offers a hygienic and convenient alternative to contact-based systems, while avoiding latent prints, pressure artifacts, and the need for specialized sensors. However, contactless images often have lower ridge clarity due to illumination variation, subcutaneous skin discoloration, and specular reflections. Flash captures preserve ridge detail but introduce noise, whereas non-flash captures reduce noise but lower ridge contrast. We propose \textbf{Fusion2Print (F2P)}, the first framework to systematically capture and fuse paired flash--non-flash contactless fingerprints. We construct a custom paired dataset, \textbf{FNF Database}\footnote{\url{https://github.com/FNFDatabase/Flash-Non-Flash-Fingerphoto-Database}}, and perform flash-non-flash subtraction to isolate ridge-preserving signals. A lightweight attention-based fusion network integrates both modalities, emphasizing informative channels and suppressing noise, followed by a U-Net enhancement module that produces an optimally weighted grayscale image. Finally, a deep embedding model with \textbf{cross-domain compatibility} is proposed that generates a discriminative and robust representation in a unified embedding space compatible with contactless, contact-based, and inked fingerprints for verification. F2P enhances ridge clarity and achieves superior recognition performance (\textbf{AUC=0.999, EER=1.12\%}) over single-capture baselines (Verifinger, DeepPrint).

\end{abstract}
    
\section{Introduction}
\label{sec:intro}

Traditional touch-based optical and capacitive sensors~\cite{Yu2023-nu} deliver high accuracy but require physical contact, causing hygiene concerns, sensor wear, latent prints, and alignment errors. Contactless fingerprint capture~\cite{Donida_Labati2019-en} avoids these issues by acquiring fingerphotos without touch, improving hygiene and usability, while avoiding the need for a specialized sensor. The COVID-19 pandemic further emphasized contactless safety. However, contactless systems still lag behind contact-based ones in accuracy and speed due to the difficulty of capturing 3D ridge patterns from 2D images under varying illumination.

A core challenge is the lack of datasets with controlled lighting. To address this, we capture consecutive flash and non-flash images with our custom-built mobile camera app that minimizes finger displacement. Flash images retain high-frequency ridge information but contain low-frequency noise; non-flash images reduce noise but lose ridge clarity. Unlike prior work relying solely on one modality~\cite{derawi}, we explore frequency-domain fusion to combine complementary signals~\cite{LengDualDCT} and preserve ridges, while suppressing noise.

We further observe that blue and green channels carry stronger ridge information, whereas the red channel is dominated by subsurface skin and blood content~\cite{G2024-uj}. This motivates an adaptive enhancement strategy that exploits channel-specific contributions. Compared to classical Gabor~\cite{Li2011-dx} and FFT-based filters, our U-Net model learns illumination-invariant, ridge-preserving representations robust to irregularities and blur.

Commercial systems such as Verifinger~\cite{neurotechnology_verifinger} and DeepPrint~\cite{Engelsma2019LearningAF} perform well on contact-based datasets but are not optimized for contactless images, showing degraded accuracy. To bridge this gap, we design a unified learning-based enhancement-embedding framework that jointly matches contact, contactless, and mixed-domain fingerprints, incorporating spatial transformations to align core regions and improve matching robustness across varied capture conditions.

Our contributions are: (i) a custom paired dataset, \textbf{Flash--Non-Flash Fingerphoto (FNF) Database}, (ii) \textbf{Fusion2Print (F2P)} an end-to-end deep learning enhancement and embedding pipeline and (iii) a unified embedding generator that improves recognition performance with robust \textbf{cross-domain matching} compared to single-capture baselines.

\section{Related Work}
\label{sec:relatedwork}

Earlier work~\cite{Sreehari2025-tz} introduced contactless fingerprint identification as a promising alternative to contact-based methods. One of the first systems appeared in 2004~\cite{song2004touchless}, motivating research to overcome contact-based limitations~\cite{kumar2018introduction}, with growing interest supported by organizations such as NIST~\cite{stanton2016usability}.

Contactless systems typically use high-resolution cameras to capture ridge, valley, and skin-texture details~\cite{Lee2006-rk}, avoiding latent prints, pressure distortions, and sensor variability common in contact-based acquisition~\cite{wang2010data}. However, they introduce challenges such as background segmentation, lower effective resolution~\cite{parziale2009advanced}, finger-pose variability, ridge-valley contrast degradation, and illumination distortions~\cite{Priesnitz2021-mq}. Early approaches relied on preprocessing-segmentation, normalization, enhancement~\cite{Chen2007-lt} using classical methods such as Gabor filtering~\cite{Erwin2019-ew}, SIFT descriptors~\cite{Kumaran2025-oy}, and core-point spatial alignment~\cite{BAHGAT201315} to achieve pose-invariant matching.

With smartphones, mobile fingerphoto acquisition became central. Systems such as~\cite{Donida_Labati2019-en,Priesnitz2022-ak} explored usability, end-to-end pipelines, and environmental robustness, while~\cite{Raghavendra2013-mn} addressed uncontrolled backgrounds, scaling, and lighting issues. Dedicated benchmarks like RidgeBase~\cite{Jawade2023-kj} and large-scale latent/fingerphoto datasets~\cite{sankaran2017learning} enabled reproducible research and stronger feature extraction.

A major challenge has been interoperability with legacy contact-based systems. Pipelines such as Contact-to-Contactless (C2CL)~\cite{Grosz2022-zx} addressed low ridge-valley contrast and pose variations, while deformation compensation methods like Robust Thin-Plate Splines (RTPS)~\cite{lin2018matching} aligned minutiae across modalities. Low-resolution scenarios were also explored~\cite{kumar2011contactless}, demonstrating feasibility for broader applications.

Deep learning has substantially advanced fingerprint recognition, with CNN- and ViT-based models~\cite{Kaplesh2024-zr} achieving state-of-the-art performance on datasets such as ISPFD~\cite{Sankaran2015-sn} and UNFIT~\cite{Chopra2018-hz}. Minutiae-aware pipelines~\cite{Tan2020-wo}
improved robustness to pose and illumination, while 3D fingerprint methods~\cite{Dong2025-jw} enabled volumetric modeling to mitigate contactless distortions. For contactless verification, siamese and attention-based architectures~\cite{8272708,9417198} have demonstrated strong cross-domain generalization by learning pairwise similarity and combining global texture with local minutiae cues, though progress remains limited by the scarcity of diverse contactless datasets for training~\cite{jcp2030036}.

Despite these advances, reliance on illumination remains a major limitation. Existing enhancement techniques largely operate on single-modality inputs. Motivated by this gap, our work introduces a paired flash--non-flash dataset and proposes an end-to-end pipeline that jointly performs enhancement and unified embedding learning. By leveraging multi-capture fusion with spatial alignment and cross-domain embedding optimization, our approach advances robustness and interoperability in contactless and contact-based fingerprint recognition.

\section{Proposed Method}
\label{sec:method}
\begin{figure}
\includegraphics[width=0.99\textwidth, trim={0 201 0 0}, clip]{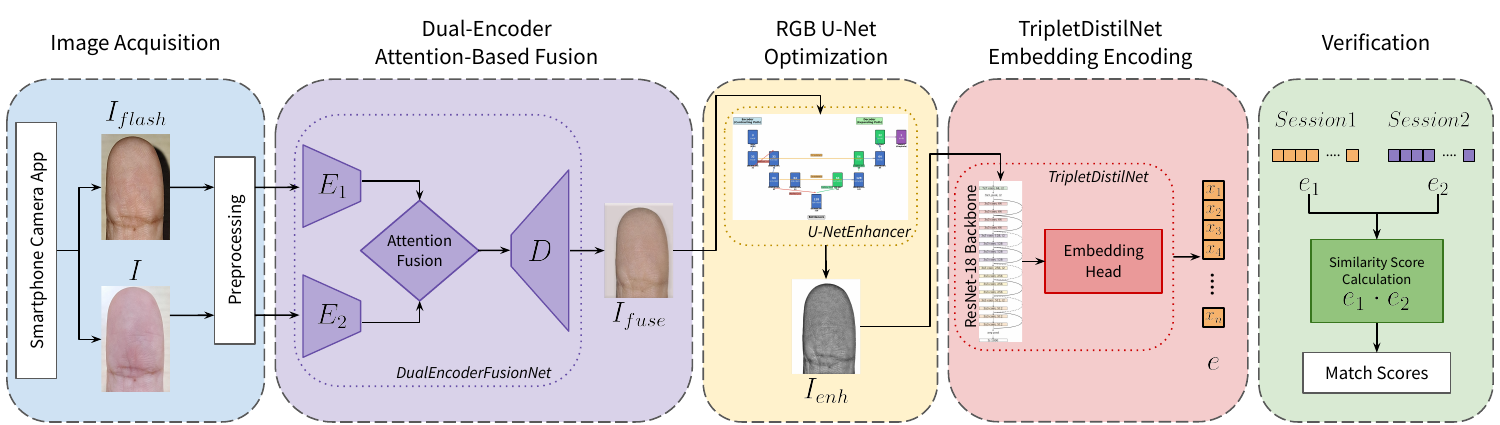}
\caption{Overview of the proposed Fusion2Print (F2P) framework. Aligned flash ($I_{\mathrm{flash}}$) and non-flash ($I$) images are processed by a dual-encoder attention fusion network, where $E_1$ and $E_2$ denote flash and non-flash encoders and $D$ produces the fused image $I_{\mathrm{fuse}}$. A U-Net--based enhancer refines $I_{\mathrm{fuse}}$ to obtain $I_{\mathrm{enh}}$, which is encoded by a ResNet-18-based feature extractor to produce discriminative embeddings. Verification is performed by comparing embeddings $e_1$ and $e_2$ from two capture sessions.}

\label{fig:2}
\end{figure}

This section presents our full contactless fingerprint enhancement and embedding framework shown in Fig.~\ref{fig:2}.

\subsection{Paired Flash--Non-Flash Capture Dataset}
\label{3.1}
We developed a cross-platform smartphone application\footnote{\url{https://github.com/FNFDatabase/Flash-Non-Flash-Fingerphoto-Database/tree/main/src/CameraApp}} using the open-source Expo Camera ~\cite{expo_camera} framework to capture paired flash and non-flash fingerprint images with minimal displacement. The app first acquires a flash image, followed by a non-flash capture after a 600 ms delay, avoiding manual flash toggling latency. All images were taken in macro mode with a fixed zoom factor of 0.45 to maintain consistent scale and ridge clarity. Using this setup on a Samsung A54, we collected a novel custom dataset, \textbf{Flash-Non-Flash Fingerphoto (FNF) Database} (see Supplementary Sections~1-2 for release details). It consists of 3,140 valid fingerprint images from 79 diverse individuals across two sessions, with 20 impressions per subject per session.

\subsection{Spectral--Domain Noise Attenuation}
\label{3.2}
Flash images contain high-frequency ridge details but are corrupted by low-frequency illumination noise; to suppress this while preserving ridge structure, we perform a flash--non-flash subtraction in the frequency domain. First, to eliminate misalignment caused by the temporal gap between captures, paired flash ($I_{\text{flash}}$) and non-flash ($I$) images are cropped to a fixed ROI and aligned using phase correlation (\textit{cv2.phaseCorrelate}), since edge-based alignment using Canny boundaries was inconsistent for truncated fingertips. Alignment translation is estimated as $(\Delta x, \Delta y) = \arg\max \mathcal{F}^{-1}\!\left\{\frac{F_1 F_2^*}{|F_1 F_2^*|}\right\}$, where $F_{1}$ and $F_{2}$ are the Fourier transforms of the paired images, and applied via \textit{cv2.warpAffine}, followed by cropping to remove border artifacts. Next, finger segmentation is performed using an $R/G$ ratio criterion ($\frac{R}{G} > T$, with $T$ empirically selected), exploiting the higher red-to-green intensity from blood concentration. Non-finger pixels are set to white to suppress background interference. For spectral attenuation, $I$ is transformed to the frequency domain and a low-pass mask $H(u,v)$ extracts low-frequency components below cutoff $f_c$ determined from the radial energy profile $R(f)$. The low-pass reconstruction of the non-flash image, $I_{\text{low}}$, is then subtracted channel-wise from the flash image,
\begin{equation}
I_{\text{diff}}^c(x,y) = I_{\text{flash}}^c(x,y) - I_{\text{low}}^c(x,y),
\end{equation}
producing a ridge-preserving differential image $I_{\text{diff}}$ with attenuated low-frequency noise. Finally, the result may be refined via grayscale conversion, CLAHE-based local contrast enhancement~\cite{Pizer2002-tf}, adaptive binarization, Gaussian smoothing, and Gabor filtering~\cite{Li2011-dx} to reinforce ridge continuity for verification(refer Supplementary Section~3 for visual results). Any remaining flash--non-flash orientation inconsistencies are mitigated later using spatial transform alignment (Sec.~\ref{sec:spatialtransform}).

\subsection{Dual-Encoder Attention-Based Fusion Network}
\label{3.3}
The proposed \textit{DualEncoderFusionNet} fuses complementary cues from paired flash ($I_\text{flash}$) and non-flash ($I$) fingerprint images to enhance ridge clarity while suppressing noise. Unlike flash-only or fixed subtraction approaches above, it learns modality-specific cues to reduce subtraction artifacts. Six-channel inputs (RGB $I_{\text{flash}}$ and $I$) are mapped to a three-channel fused output $I_{\text{fuse}}$. The network uses a dual-encoder fusion-decoder structure with perceptual and frequency-based enhancement losses as follows:

\paragraph{Dual Encoders.} Two parallel convolutional encoders extract hierarchical ridge and illumination features from $I_{\text{flash}}$ and $I$, for each layer $F_{x}=\text{ReLU}(\text{BN}(\text{Conv}(x)))$. 

\paragraph{Attention-Based Feature Fusion.} Features are merged by pixel-wise soft attention with spatial weights $W_1,W_2$ ($W_1+W_2=1$): $F_{\text{fuse}} = W_1\odot F_{\text{flash}} + W_2\odot F$. 

\paragraph{Frequency and Edge-Aware Decoder.} The decoder, $D$, reconstructs $I_{\text{fuse}}$ with ridge-preserving enhancement. High-frequency components are selectively amplified via:
$
I' = D(F_{\text{fuse}})\!\cdot\!\Big(1 + \lambda \!\cdot\! \frac{\log(1 + |\mathcal{F}\{D(F_{\text{fuse}})\}|)}{\max(|\mathcal{F}\{D(F_{\text{fuse}})\}|) + \epsilon}\Big),
$
where $\mathcal{F}$ denotes the Fourier transform and $\lambda$ controls amplification. Edge refinement uses a residual convolution:
$I_{\text{fuse}} = I' + \beta \cdot \text{Conv}_{\text{edge}}(I').$
A segmentation mask $M$ suppresses background regions during loss computation (see Sec.~\ref{3.2} for mask generation). 

\subsubsection{Training Strategy and Optimization.} Training introduces the difference image $I_{\text{diff}}$ as supervision, improving ridge definition over standard CNN baselines. Inputs are padded and resized to $512\times512$, losses are applied only within $M$, and the model is trained with Adam optimizer(lr $=10^{-4}$, weight decay $=10^{-5}$) for 30 epochs using a 70-15-15 split on our flash--non-flash dataset. A learning rate scheduler stabilizes convergence, and a custom loss drives reconstruction while rewarding high-frequency details to achieve finer detail than the target.

\subsubsection{Custom Combined Loss.} A composite loss objective integrates pixel, structural, frequency, edge, and perceptual constraints, to surpass the target $I_{\text{diff}}$. L1 loss ensures spatial fidelity: $\mathcal{L}_{L1}=\frac{1}{N}\sum_i \left| {I}_{\text{fuse},i} - I_{\text{diff},i} \right|$. SSIM maintains structural consistency: $\mathcal{L}_{SSIM}=1-\text{SSIM}({I}_{\text{fuse}}, I_{\text{diff}})$. Fourier loss rewards high-frequency texture: $\mathcal{L}_{Fourier}=\max\left(0,\, E_{hf}(I_{\text{diff}}) - E_{hf}({I}_{\text{fuse}})\right)$. Edge loss sharpens ridge gradients:  $\mathcal{L}_{Edge} = 
\|\nabla_x({I}_{\text{fuse}}) - \nabla_x(I_{\text{diff}})\|_1 +
\|\nabla_y({I}_{\text{fuse}}) - \nabla_y(I_{\text{diff}})\|_1$. Perceptual loss matches feature-space similarity via VGG16\cite{simonyan2015deepconvolutionalnetworkslargescale}: $\mathcal{L}_{Perceptual} = 
\| \phi({I}_{\text{fuse}}) - \phi(I_{\text{diff}}) \|_1$, (refer Supplementary Section~4). The total empirically weighted loss, with $\alpha=0.6$, $\beta=0.1$, $\gamma=0.15$, $\delta=0.1$, and $\epsilon=0.05$, is
\begin{equation}
\mathcal{L}_{total}=\alpha\mathcal{L}_{L1}+\beta\mathcal{L}_{SSIM}+\gamma\mathcal{L}_{Fourier}+\delta\mathcal{L}_{Edge}+\epsilon\mathcal{L}_{Perceptual}
\end{equation}

\subsection{Color-Space Ridge Enhancement}

To maximize ridge clarity while minimizing light and skin reflectance artifacts, we design a color-space enhancement pipeline combining channel-wise RGB contrast analysis and a U-Net-based enhancement model.

\subsubsection{Channel-Wise RGB Contrast Analysis.}
We assess each RGB channel’s contribution to ridge visibility in flash $(I_{\text{flash}})$ and non-flash $(I)$ images(see Supplementary Section~5.1) using local contrast, defined as the mean block-wise standard deviation:
$
C_{\text{local}}=\frac{1}{N}\sum_{k=1}^{N}\text{std}(B_k)
$
with $B_k$ the $k^{\text{th}}$ non-overlapping block. Across channels we observed:
\[
\text{Blue}>\text{Green}>\text{Red}.
\]

\vspace{-2em}
\begin{figure}[!ht]
\centering
\includegraphics[width=0.59\linewidth]{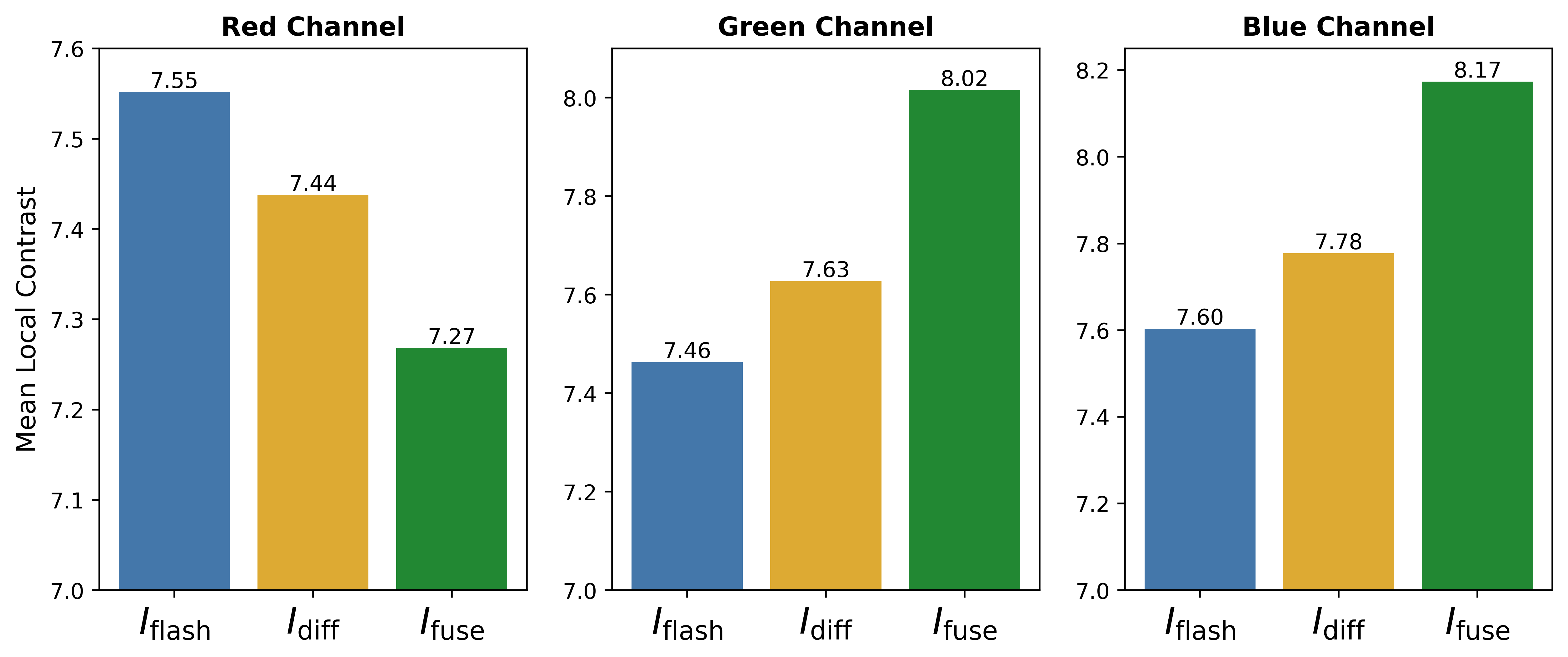}
\caption{Channel-wise RGB local contrast of $I_{\text{flash}}$, $I_{\text{diff}}$, and $I_{\text{fuse}}$.}
\label{fig:3}
\end{figure}
\vspace{-1em}

Fig.~\ref{fig:3} shows that  
$C_{\text{local}}(I_{\text{fuse}}^{B,G})>C_{\text{local}}(I_{\text{diff}}^{B,G})>C_{\text{local}}(I_{\text{flash}}^{B,G})$, indicating stronger ridge detail in blue/green bands, whereas  
$C_{\text{local}}(I_{\text{fuse}}^{R})<C_{\text{local}}(I_{\text{diff}}^{R})<C_{\text{local}}(I_{\text{flash}}^{R})$, reflecting weaker ridge information in red. We observe, blue and green channels are emphasized during frequency enhancement, while red is deweighted to suppress skin and subsurface-scattering artifacts. Based on this, we build:

\subsubsection{U-Net Enhancement Model.}
\label{3.4}
\textit{U-NetEnhancer} $(\mathcal{U}_\theta)$ maps the RGB fused image $(I_{\text{fuse}})$ to an enhanced custom weighted grayscale ridge map$(I_{\text{enh}})$ to give more weight to ridge-rich blue and green channels $(I_{\text{enh}}=\mathcal{U}_\theta(I_{\text{fuse}}))$. Each encoder--decoder block contains two Conv--BN--ReLU layers with skip connections, and a sigmoid output normalizes intensity. Unsupervised variants were overly smooth, so we use supervised training with aligned flash grayscale $(I_{\text{flash}})$ and a Gabor-based ridge loss. Inputs are resized and black-padded to $512\times512$ and losses are computed only within the fingerprint mask, $M$. Training uses a 70--15--15 split on our $I_{\text{fuse}}$ dataset, Adam Optimizer ($lr=10^{-4}$), 30 epochs, and Gaussian perturbations for robustness.

\textbf{Empirically Weighted Custom Loss:}
\begin{equation}
\mathcal{L}_{\text{final}}=
\alpha\mathcal{L}_{L1}
-\beta\mathcal{R}_{\text{contrast}}
+\gamma\mathcal{L}_{SSIM}
+\delta\mathcal{L}_{Gabor}
+\eta\mathcal{L}_{edge},
\end{equation}

where,\\
\noindent
\begin{minipage}{0.48\linewidth}
\begin{equation}
\mathcal{L}_{L1}=
\frac{\sum |I_{\text{enh}}-I_{\text{flash}}|M}{\sum M+\epsilon}, \alpha=0.4
\end{equation}
\end{minipage}
\hfill
\begin{minipage}{0.48\linewidth}
\begin{equation}
\mathcal{R}_{\text{contrast}}=
\frac{1}{N}\sum_{p}\sigma_p(I_{\text{enh}}) w_p, \beta=0.2
\end{equation}
\end{minipage}

\noindent
\begin{minipage}{0.48\linewidth}
\begin{equation}
\mathcal{L}_{SSIM}=1-\text{SSIM}(I_{\text{enh}},I_{\text{flash}}), \gamma=0.3
\end{equation}
\end{minipage}
\begin{minipage}{0.48\linewidth}
\begin{equation}
\mathcal{L}_{edge}=\|\nabla_{\text{Sobel}}I_{\text{enh}}-\nabla_{\text{Sobel}}I_{\text{flash}}\|_1,\delta=0.1
\end{equation}
\end{minipage}

\begin{equation}
\mathcal{L}_{Gabor}=\frac{\sum |I_{\text{enh}}-G_{\text{ref}}|W_{\text{ridge/valley}}M}{\sum M+\epsilon},\eta=0.1
\end{equation}

Here $\sigma_p$ is patch-wise standard deviation, $w_p$ are adaptive channel weights $(w_B>w_G>w_R)$. While $\mathcal{L}_{L1}$ and $\mathcal{L}_{SSIM}$ enable reconstruction, adding $\mathcal{R}_{\text{contrast}}$, $\mathcal{L}_{Gabor}$, and $\mathcal{L}_{edge}$ enhances ridge--valley contrast, continuity, and sharpness---exceeding the quality of grayscale $I_{\text{flash}}$, (see Supplementary Section~5.2, 5.3). The final $I_{\text{enh}}$ is further refined using Gabor filtering (Sec.~\ref{3.2}).

\subsection{Deep Embedding Learning for Verification}
We evaluate baseline verification and propose a deep embedding model to learn discriminative representations from enhanced ridge maps.

\subsubsection{Baseline Verification.}
Fingerprint verification relies on \textit{minutiae-} or \textit{texture-based} cues. To assess enhancement quality, Session~1 fingerprints are enrolled and matched with Session~2 using:

\begin{itemize}
    \item \textbf{VeriFinger}~\cite{neurotechnology_verifinger}: a commercial minutiae-based matcher using 20--40 ridge endings/bifurcations per pair, achieving FAR@0.01\%.
    \item \textbf{DeepPrint}~\cite{Engelsma2019LearningAF}: an embedding-based dual-branch CNN producing a 512-D embedding $\mathbf{f}$, with similarity via cosine:
    $S(\mathbf{f}_1,\mathbf{f}_2)=\frac{\mathbf{f}_1\cdot\mathbf{f}_2}{\|\mathbf{f}_1\|_2\,\|\mathbf{f}_2\|_2}$.
\end{itemize}

Metrics reported include ROC--AUC(overall discriminative ability), False Accept Rate (FAR), False Reject Rate (FRR), and Equal Error Rate (EER),(see Supplementary Section~6.1 for more information).

\subsubsection{Proposed Embedding Model.}
\label{3.5}
We introduce \textit{TripletDistilNet}(see Supplementary Sections~6.2 for training variants), combining triplet metric learning with knowledge distillation from DeepPrint. Earlier Siamese and LBP-based variants showed weaker separability.

\medskip
\noindent\textbf{Triplet Dataset.}
Each subject provides paired contactless fingerprints $(I_i^{(1)},I_i^{(2)})$. Triplets $(I_a,I_p,I_n)$ satisfy  
$I_a,I_p$ share identity and $I_n$ differs. Six semi-hard triplets per identity ensure balanced difficulty.

\medskip
\noindent\textbf{Embedding Network.}
A pretrained ResNet-18~\cite{He_2016_CVPR} backbone and embedding head is fine-tuned on enhanced fingerprints $(I_{\text{enh}})$ with geometric and Gaussian-noise augmentation. Images map to $\ell_2$-normalized embeddings: 
$\mathbf{e}=\frac{f_\theta(x)}{\|f_\theta(x)\|_2}
\in 
\mathbb{R}^{128}\,(256{\times}256),\ 
\mathbb{R}^{256}\,(512{\times}512).
$

\medskip
\noindent\textbf{Loss Formulation.}
Cosine-distance triplet loss with margin $m$ drives discriminative learning:
\begin{equation}
\mathcal{L}_{\text{triplet}}=
\max\!\big(0,\ d(\mathbf{e}_a,\mathbf{e}_p)-d(\mathbf{e}_a,\mathbf{e}_n)+m\big),
\end{equation}
with semi-hard constraint  
$d(\mathbf{e}_a,\mathbf{e}_p)<d(\mathbf{e}_a,\mathbf{e}_n)<d(\mathbf{e}_a,\mathbf{e}_p)+m(=0.3)$.
DeepPrint teacher embeddings $(\mathbf{t}_a,\mathbf{t}_p,\mathbf{t}_n)$ are projected to student dimensionality and used for cosine-based distillation:
\begin{equation}
\mathcal{L}_{\text{distill}}=
1-\tfrac{1}{3}\!\left[
\cos(\mathbf{e}_a,\mathbf{t}_a)+
\cos(\mathbf{e}_p,\mathbf{t}_p)+
\cos(\mathbf{e}_n,\mathbf{t}_n)
\right].
\end{equation}
The combined objective is
\begin{equation}
\mathcal{L}_{\text{total}}=
\mathcal{L}_{\text{triplet}}
+\alpha\,\mathcal{L}_{\text{distill}},\quad
\alpha=0.7.
\end{equation}
Training uses AdamW optimizer($lr=10^{-4}$, batch size 16) with learning rate scheduling and per-epoch triplet re-sampling.

\medskip
\noindent\textbf{Contact Fine-tuning.}
To unify embeddings across contact, contactless, and mixed-domain inputs, we finetune the \textit{TripletDistilNet} head and a few backbone layers on contact fingerprints (Supplementary Sec.~8.2).

\medskip
\noindent\textbf{Verification.}
Mated/non-mated pairs are classified using a threshold on cosine similarity between \textit{TripletDistilNet} embeddings, with identity-wise performance reported.

\subsection{End-to-End Enhancement--Embedding Pipeline}
\label{3.6}
Building on the frequency-aware fusion, RGB enhancement, and deep embedding modules, we integrate them into a unified framework termed \textbf{Fusion2Print (F2P)}. The full pipeline jointly processes paired flash and non-flash fingerprints to produce discriminative embeddings for verification.

\subsubsection{Architecture Overview.}
F2P composes three pretrained modules in sequence:
\begin{itemize}
    \item \textit{DualEncoderFusionNet}: frequency- and edge-aware fusion preserving ridge continuity,
    \item \textit{U-NetEnhancer}: channel-balanced contrast correction with blue--green emphasis,
    \item \textit{TripletDistilNet}: metric-learning backbone trained on enhanced fingerprints.
\end{itemize}

Given paired inputs $I$, $I_{\text{flash}}$ and fingerprint mask $M$, the $\ell_2$-normalized embedding $\mathbf{e}$ is:
\begin{equation}
I, I_{\text{flash}}
\xrightarrow{\mathcal{F}_{\text{fusion}}}
I_{\text{fuse}}
\xrightarrow{\mathcal{U}_{\text{enh}}}
I_{\text{enh}}
\xrightarrow{\mathcal{E}_{\text{triplet}}}
\mathbf{e} \in \mathbb{R}^{d}, \; d \in \{128,256\}.
\end{equation}

\subsubsection{Fine-tuning.}
End-to-end fine-tuning improves cross-session consistency and discrimination.  
\textit{DualEncoderFusionNet} and \textit{U-NetEnhancer} modules are partially frozen, while the embedding head is trained on paired flash--non-flash samples (80/20 split). Training uses Adam optimizer($\text{lr}=10^{-6}$, weight decay $10^{-7}$) for 10 epochs with gradient clipping and cosine-based supervision.

The loss combines a soft-margin triplet term with cosine alignment:
\begin{equation}
\mathcal{L}_{\text{fine}} =
w_t \log\!\big(1 + \exp(d_{ap} - d_{an} + \delta)\big)
+ w_i \big(1 - \cos(\mathbf{e}_a, \mathbf{e}_p)\big),
\end{equation}
where $d_{ap}$ and $d_{an}$ are anchor--positive and anchor--negative distances, $\delta = 0.08$, and $(w_t, w_i) = (0.75, 0.25)$ and the function controls the trade-off between separation and identity coherence, (see Supplementary Section~7).

\subsubsection{Orientation Alignment and Centering via Core Point Detection.}
\label{sec:spatialtransform}

To ensure geometric consistency across sessions, flash and non-flash images are upright-aligned and centered before entering F2P.
The rotation angle is estimated from boundary slopes:
\begin{equation}
\theta =
\tfrac{1}{2}\big(\arctan(m_{\text{left}}) + \arctan(m_{\text{right}})\big),
\end{equation}
and each image is rotated by $-\theta$.

The core point $(x_c, y_c)$ ~\cite{BAHGAT201315} is then detected from the orientation field using coherence and Poincaré index:
\begin{equation}
C_{ij} = 
\frac{\sqrt{V_x^2 + V_y^2}}
{\sum (G_x^2 + G_y^2) + \epsilon},
\qquad
P_{ij} \approx +0.5.
\end{equation}
Each fingerprint is translated so $(x_c, y_c)$ aligns with the image center, yielding a stable ROI for fine-tuning (see Supplementary Section~7.1 for visual results).
This spatial normalization supports robust end-to-end optimization of the enhancement-embedding pipeline.

\section{Experiments \& Results}
\label{sec:exp}
 Since no other public paired flash--non-flash fingerprint dataset exists, we evaluate the proposed F2P pipeline on our curated dataset and compare against multiple settings:
non-flash input ($I$), flash input ($I_{\text{flash}}$), manual subtraction output($I_{\text{diff}}$), fused output ($I_{\text{fuse}}$), and final enhanced image ($I_{\text{enh}}$).

\subsection{Ridge Structure Quality Analysis}
We quantitatively assess ridge quality scores using NFIQ2~\cite{nfiq2} and three complementary metrics, as NFIQ2 is not trained for contactless fingerprints:

\begin{itemize}
\item \textbf{Local Contrast} captures patch-wise intensity variation indicating ridge-valley separation. 
\begin{figure}
  \centering
   \includegraphics[width=0.8\linewidth]{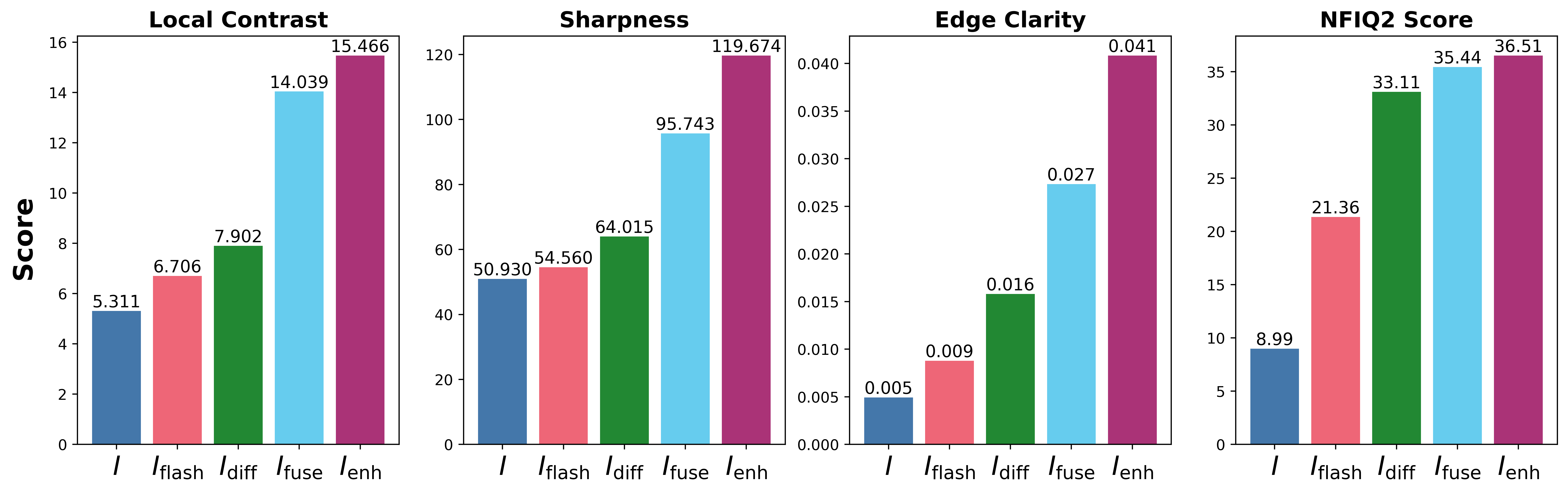}
   \caption{Comparison of ridge-quality scores across input types.}
   \label{fig:4}
\end{figure}
\vspace{-1em}
\item \textbf{Sharpness} computes variance of the Laplacian capturing high-frequency ridge detail. 
\item  \textbf{Edge Clarity} measures proportion of Canny-detected edges reflecting ridge boundary visibility.
\end{itemize}
We see that $I_{\text{enh}}$ outperforms $I_{\text{fuse}}$, $I_{\text{diff}}$, $I_{\text{flash}}$ and $I$ across all the metrics in Fig.~\ref{fig:4}.


\subsection{Baseline Matcher-Based Verification Results}
We first performed baseline verification performance for $I$, $I_{\text{flash}}$, $I_{\text{diff}}$, $I_{\text{fuse}}$, $I_{\text{enh}}$ and $I_{\text{enh\_ST}}$ (data with spatial transform on $I_{\text{enh}}$) datasets using DeepPrint (embedding-based) and VeriFinger(minutiae-based) matchers in Tab.~\ref{tab:1}. We see that our enhancement processing improves matching accuracy.

\vspace{-2em}
\begin{table}
\centering
\caption{Verification performance of DeepPrint and VeriFinger across our dataset variations.}
\label{tab:1}
\setlength{\tabcolsep}{10pt}
\resizebox{\textwidth}{!}{%
\begin{tabular}{lccccccc}
\hline
\textbf{Matcher} & \textbf{Metric} & $I$ & $I_{\text{flash}}$ & $I_{\text{diff}}$ & $I_{\text{fuse}}$ & $I_{\text{enh}}$ & \textbf{$I_{\text{enh\_ST}}$} \\
\hline
\multirow{2}{*}{VeriFinger} 
    & AUC & 0.69 & 0.81 & 0.86 & 0.90 & 0.93 & \textbf{0.94} \\ 
    & EER (\%) & 30.90 & 19.11 & 14.23 & 11.30 & 8.05 & \textbf{7.60} \\ 
\hline
\multirow{2}{*}{DeepPrint} 
    & AUC & 0.64 & 0.74 & 0.87 & 0.88 & 0.94 & \textbf{0.97} \\ 
    & EER (\%) & 37.45 & 33.06 & 21.67 & 19.75 & 12.68 & \textbf{8.92} \\ 
\hline
\end{tabular}
}
\end{table}
\vspace{-2em}

\subsection{Ablation Study on TripletDistilNet}

We evaluate \textit{TripletDistilNet} verification performance across all enhancement variants (\(I\), \(I_{\text{flash}}\), \(I_{\text{diff}}\), \(I_{\text{fuse}}\), \(I_{\text{enh}}\)) as summarized in Tab.~\ref{tab:2}. All models use 128-D embeddings for efficiency, and each configuration measures how training on one domain generalizes to others, capturing cross-modality consistency.

\begin{table}[!ht]
\centering
\caption{Verification performance across different train--test configurations for the \textit{TripletDistilNet} embedding module.}
\label{tab:2}
\renewcommand{\arraystretch}{1.2}
\resizebox{\textwidth}{!}{%
\begin{tabular}{lcccccccccc}
\hline
\multicolumn{1}{c}{\textbf{Test Dataset $\rightarrow$}} 
& \multicolumn{10}{c}{} \\[-3pt]
\multicolumn{1}{c}{\textbf{Train Config $\downarrow$}} 
& \multicolumn{2}{c}{$I$} 
& \multicolumn{2}{c}{$I_{\text{flash}}$} 
& \multicolumn{2}{c}{$I_{\text{diff}}$} 
& \multicolumn{2}{c}{$I_{\text{fuse}}$} 
& \multicolumn{2}{c}{$I_{\text{enh}}$} \\
\hline
 & AUC & EER (\%) & AUC & EER (\%) & AUC & EER (\%) & AUC & EER (\%) & AUC & EER (\%) \\
\hline
$I$                     & \textbf{0.91} & \textbf{16.31} & 0.80 & 26.31 & 0.82 & 24.71 & 0.86 & 21.53 & 0.91 & 15.97 \\
$I_{\text{flash}}$      & 0.87 & 20.64 & 0.969 & 9.49 & 0.973 & 7.77 & 0.989 & 4.71 & 0.994 & 3.44 \\
$I_{\text{diff}}$       & 0.86 & 21.21 & 0.955 & 10.96 & 0.975 & 8.66 & 0.984 & 5.99 & 0.996 & 2.74 \\
$I_{\text{fuse}}$       & 0.88 & 20.83 & 0.96 & 9.43 & 0.98 & 7.20 & 0.986 & 5.16 & 0.995 & 2.68 \\
\textbf{$I_{\text{enh}}$}       & 0.874 & 20.57 & \textbf{0.979} & \textbf{7.39} & \textbf{0.989} & \textbf{5.10} & \textbf{0.992} & \textbf{4.71} & \textbf{0.997} & \textbf{2.29} \\
\hline
\end{tabular}
}
\end{table}

\begin{figure}[!ht]
  \centering
  \begin{subfigure}{0.28\linewidth}
    \includegraphics[width=0.99\linewidth]{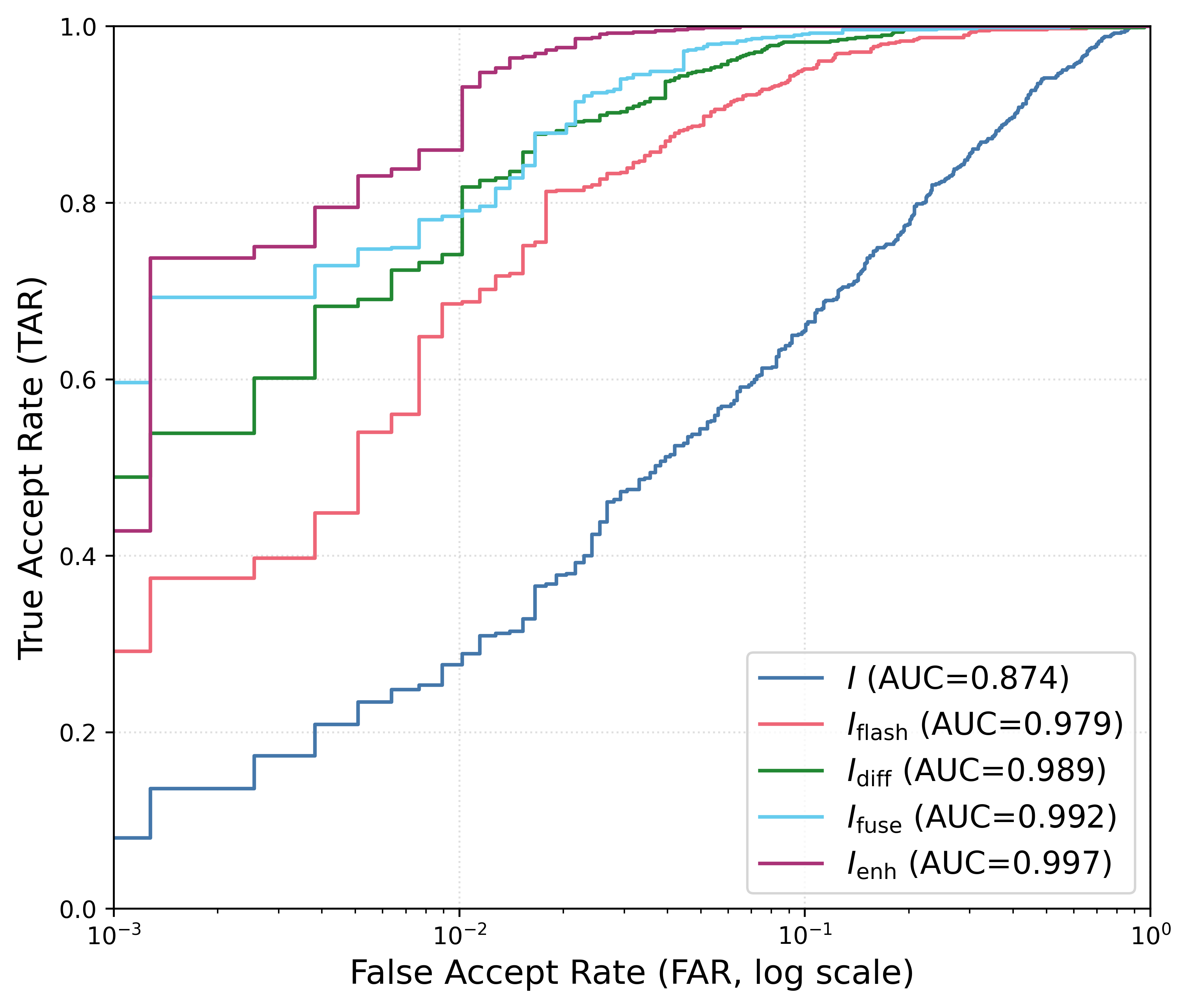}
    \caption{ROC Curve}
    \label{fig:5a}
  \end{subfigure}
  \hspace{0.9em}
  \begin{subfigure}{0.28\linewidth}
    \includegraphics[width=0.99\linewidth]{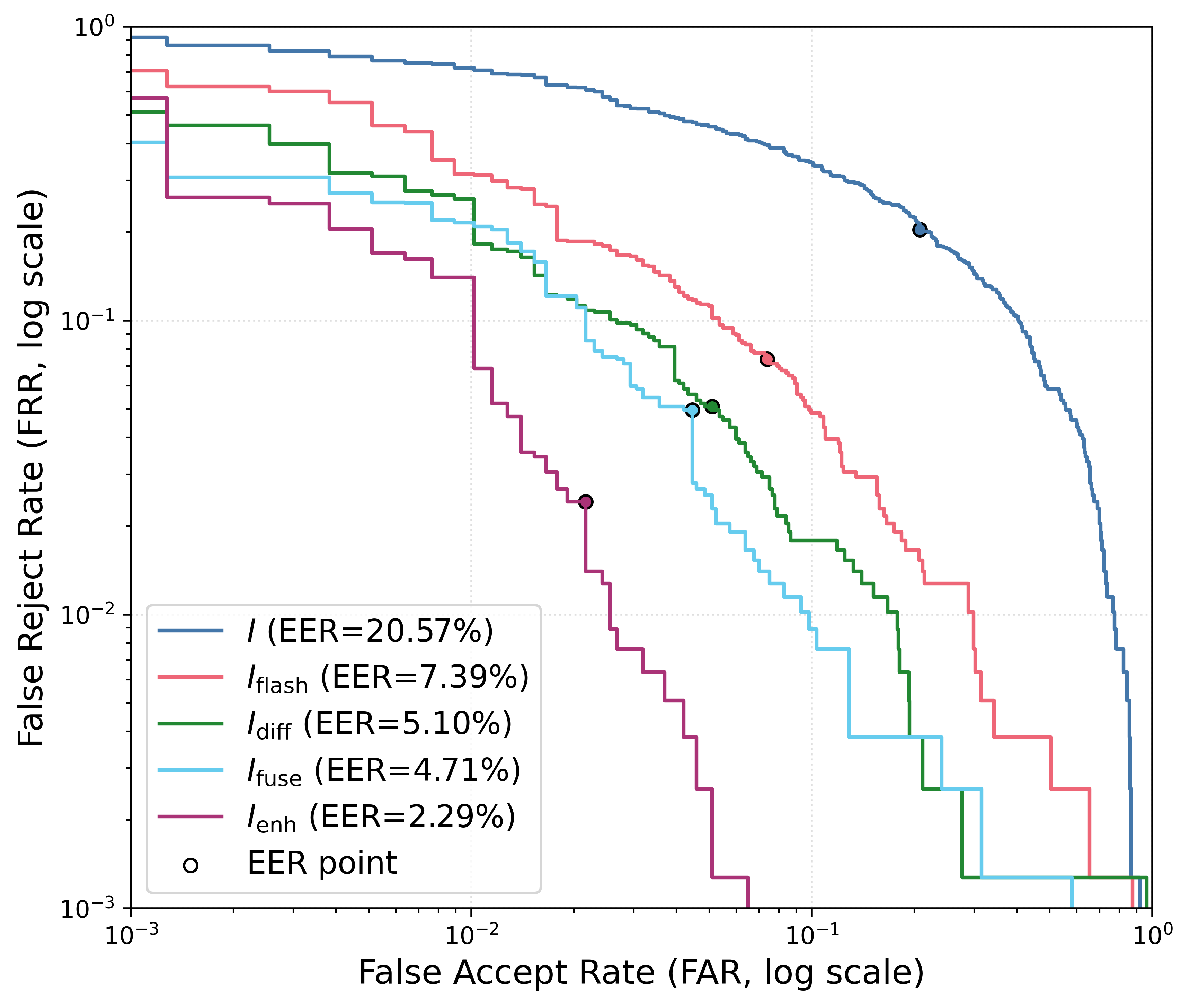}
    \caption{FRRvsFAR Curve}
    \label{fig:5b}
  \end{subfigure}
  \caption{ROC and FRRvsFAR curves for the best \textit{TripletDistilNet} model trained on $I_{\text{enh}}$.}
  \label{fig:5}
\end{figure}

Performance improves consistently with stronger enhancement: training on \(I_{\text{enh}}\) achieves the best AUC (0.997), outperforming both flash (0.969) and non-flash (0.91) baselines. The corresponding ROC and FRR--FAR curves for the top-performing configuration (trained on \(I_{\text{enh}}\)) are shown in Fig.~\ref{fig:5}.

\begin{figure}[!ht]
  \centering
  \begin{subfigure}{0.46\linewidth}
    \includegraphics[width=0.99\linewidth]{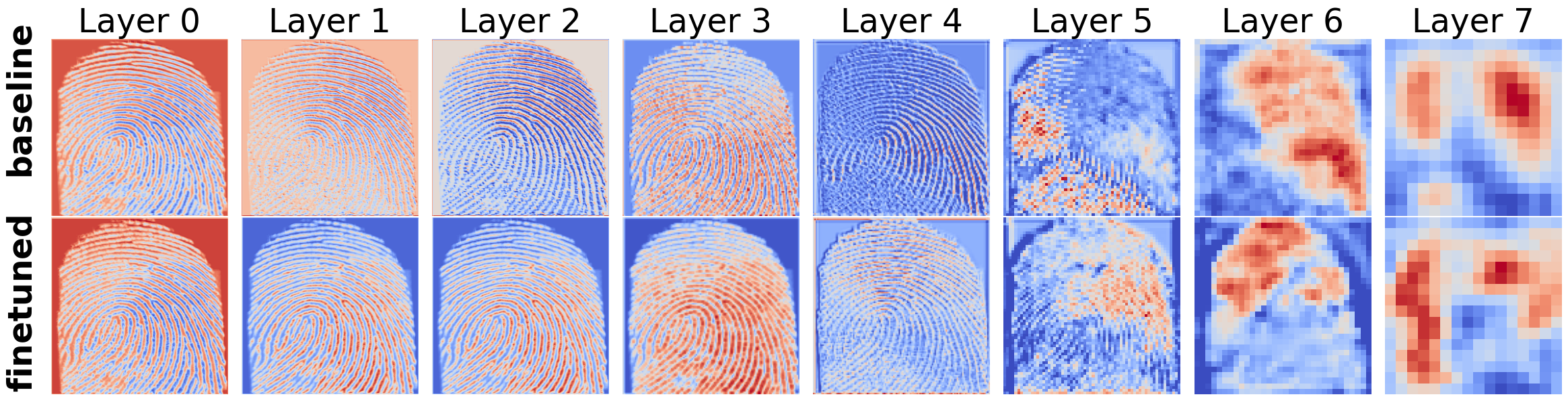}
    \caption{Contact-based Sample}
    \label{fig:6a}
  \end{subfigure}
  \hspace{0.1em}
  \begin{subfigure}{0.51\linewidth}
    \includegraphics[width=0.99\linewidth]{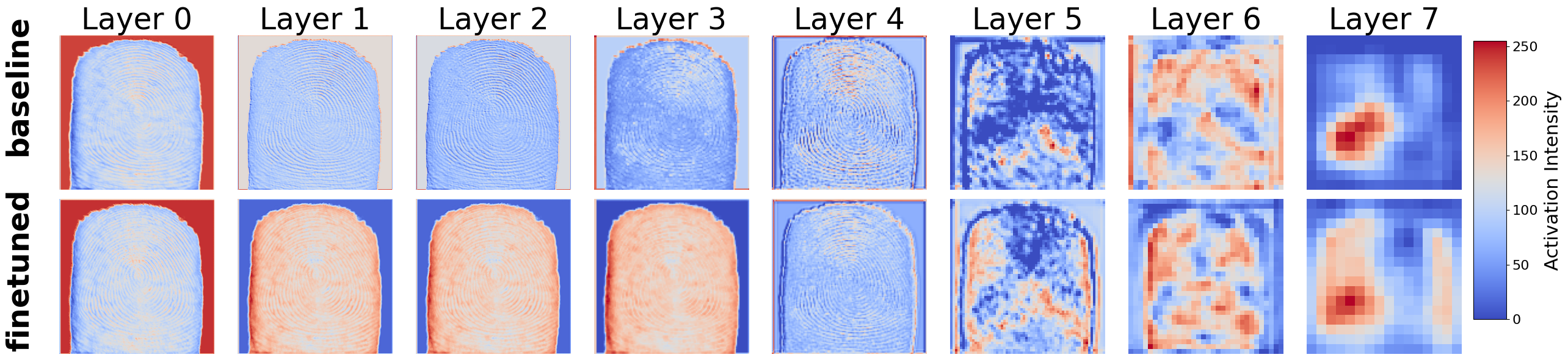}
    \caption{Contactless Sample}
    \label{fig:6b}
  \end{subfigure}
  \caption{Activation maps of \textit{TripletDistilNet} before and after fine-tuning, highlighting higher ridge- and core-focused attention (red) across contact and contactless domains.}
  \label{fig:6}
\end{figure}

As seen in Tab.~\ref{tab:3}, \textit{TripletDistilNet} trained on \(I_{\text{enh}}\) exhibits strong cross-dataset performance but weaker contact-domain accuracy. Fine-tuning the backbone on contact data significantly improves contact performance while preserving contactless discriminability. Activation-map statistics further indicate stable adaptation, especially in early layers, without embedding drift, supporting improved cross-domain generalization (see Fig.~\ref{fig:6} and Supplementary Section~8).

A mixed-domain setup was also evaluated, but its averaged cross-dataset performance gave limited clarity on dataset-specific performance.

\vspace{-2em}
\begin{table}[!ht]
\centering
\caption{Cross-dataset verification performance of DeepPrint and \textit{TripletDistilNet} (with and without contact fine-tuning). Datasets include our FNF database; contact (HKPolyU-C~\cite{lin2018matching}, SP~\cite{fvs2002}) and contactless (HKPolyU-CL~\cite{lin2018matching}, UNFIT~\cite{Chopra2018-hz}) benchmarks; and the mixed-domain dataset HKPolyU-CPhoto (see Supplementary Sec.~8 for further details and cross-device experiments).}
\label{tab:3}
\setlength{\tabcolsep}{2pt}
\resizebox{\textwidth}{!}{%
\begin{tabular}{lccccccc}
\hline
\textbf{Method} & \textbf{Metric} 
& \textbf{FNF (Ours)} 
& \textbf{HKPolyU-CL} 
& \textbf{UNFIT}  
& \textbf{HKPolyU-C} 
& \textbf{HKPolyU-CPhoto} 
& \textbf{SP} \\
\hline
\multirow{2}{*}{DeepPrint}
    & AUC       & 0.94  & 0.92  & 0.90 & 0.97 & 0.96  & 0.97   \\
    & EER (\%)  & 12.68 & 15.87 & 20.38 & 8.63  & 9.08 & 10.6   \\
\hline
\multirow{2}{*}{TripletDistilNet}
    & AUC       & \textbf{0.997} & 0.96 & 0.95  & 0.91 & 0.83 & 0.89 \\
    & EER (\%)  & \textbf{2.29}  & 10.77 & 11.14 & 15.98 & 24.20 & 19.35 \\
\hline
\multirow{2}{*}{TripletDistilNet w/ Contact FT}
    & AUC       & 0.994 & \textbf{0.962} & \textbf{0.97} & \textbf{0.988} & \textbf{0.975} & \textbf{0.982} \\
    & EER (\%)  & 3.31  & \textbf{10.42}  & \textbf{9.86} & \textbf{4.74}  & \textbf{9.25}  & \textbf{4.86} \\
\hline

\end{tabular}
}
\end{table}
\vspace{-2em}

\subsection{Verification Performance of Fusion2Print}
\begin{figure}
  \centering
  \begin{subfigure}{0.28\linewidth}
    \includegraphics[width=0.99\linewidth]{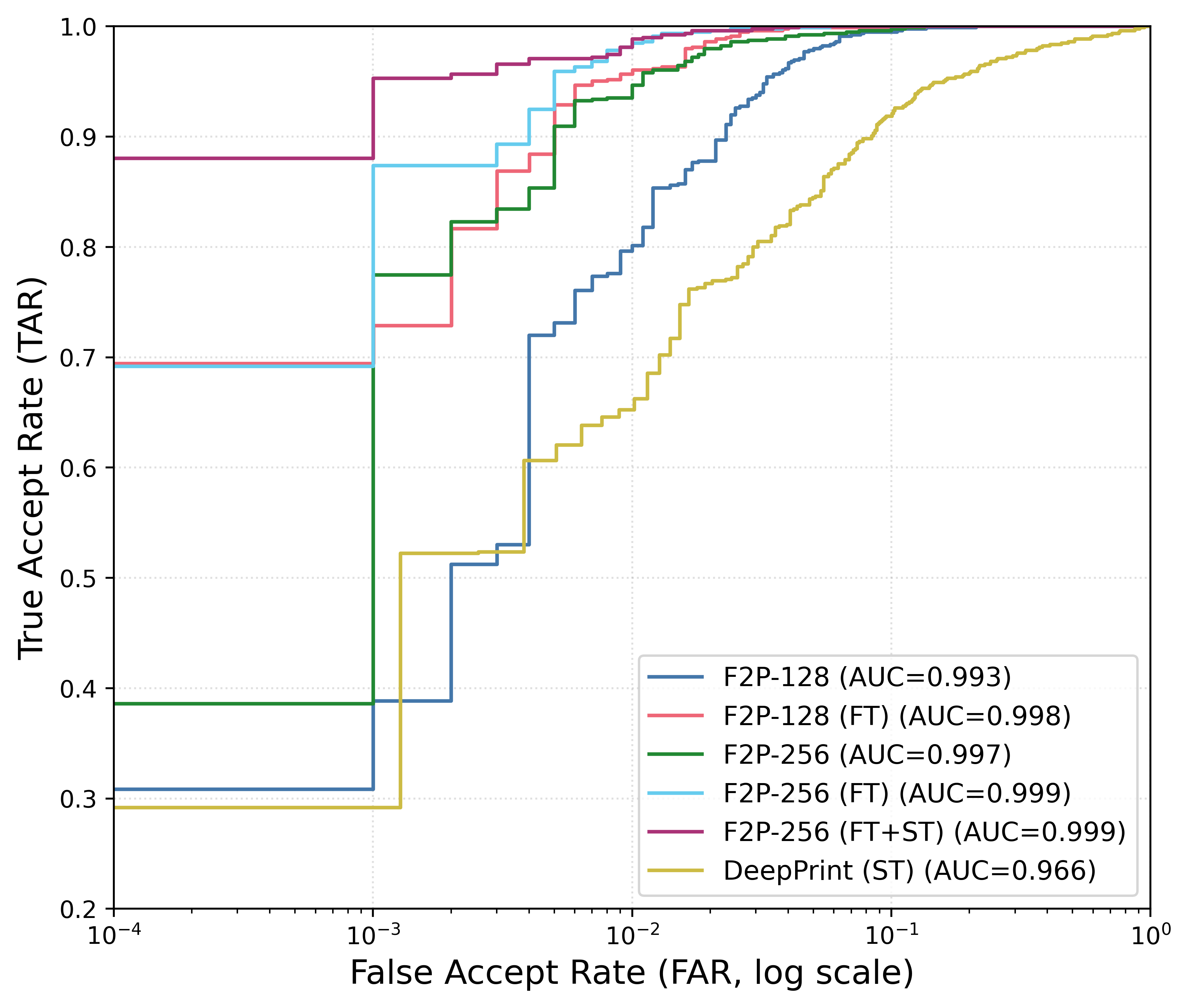}
    \caption{ROC Curve}
    \label{fig:7a}
  \end{subfigure}
  \hspace{0.9em}
  \begin{subfigure}{0.28\linewidth}
    \includegraphics[width=0.99\linewidth]{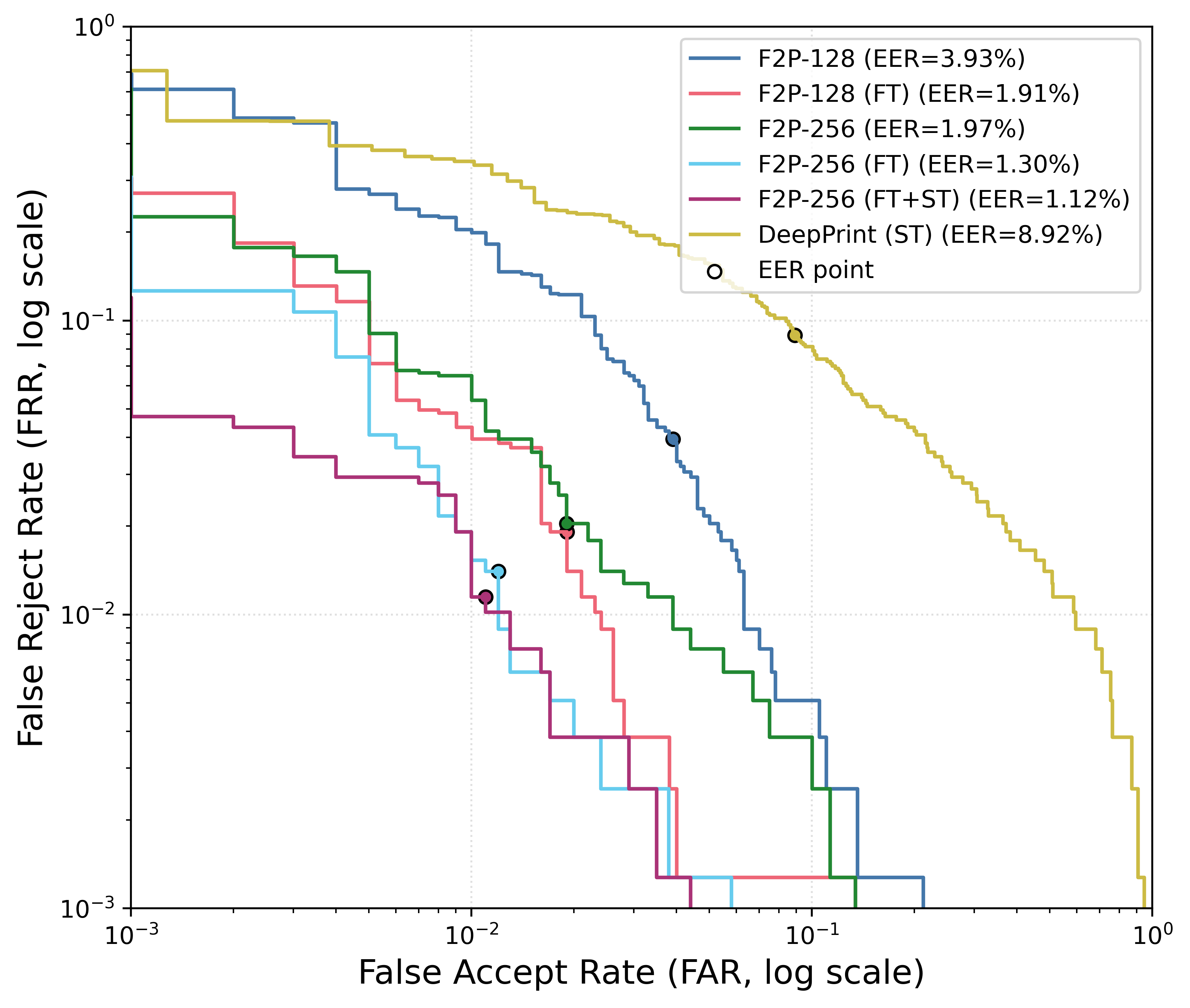}
    \caption{FRRvsFAR Curve}
    \label{fig:7b}
  \end{subfigure}
  \caption{ROC and FRRvsFAR curves for DeepPrint baseline and Fusion2Print pipeline variants.}
  \label{fig:7}
\end{figure}

We evaluated the verification performance of embeddings generated by the complete Fusion2Print pipeline for both 128- and 256-dimensional representations. The average templating latency was \textit{0.28s} per identity for  Apple M4 GPU and \textit{0.34s} for a Samsung A54 device. We compared the results of the baseline Fusion2Print model and its fine-tuned variants to assess performance gains. Observing improved results with 256-dimensional embeddings, we further fine-tuned the pipeline using spatially transformed data to enhance robustness. The outcomes are summarized in Tab.~\ref{tab:4} and Tab.~\ref{tab:5} and visualized in Fig.~\ref{fig:7}.

\begin{figure}[!ht]
  \centering



  \begin{subfigure}{0.24\linewidth}
    \includegraphics[width=\linewidth]{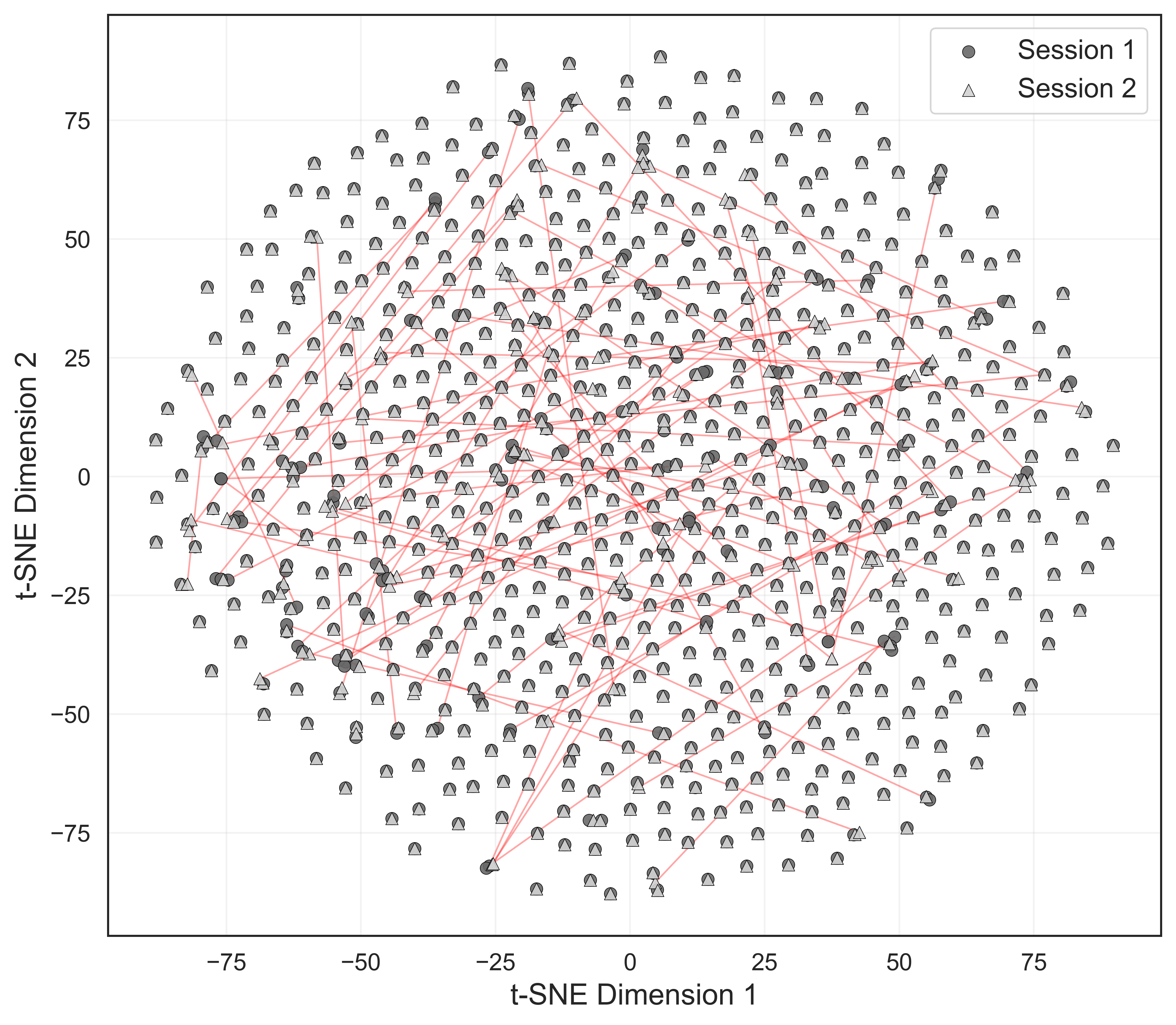}
    \caption{DeepPrint}
    \label{fig:8a}
  \end{subfigure}
  \hfill
  \begin{subfigure}{0.24\linewidth}
    \includegraphics[width=\linewidth]{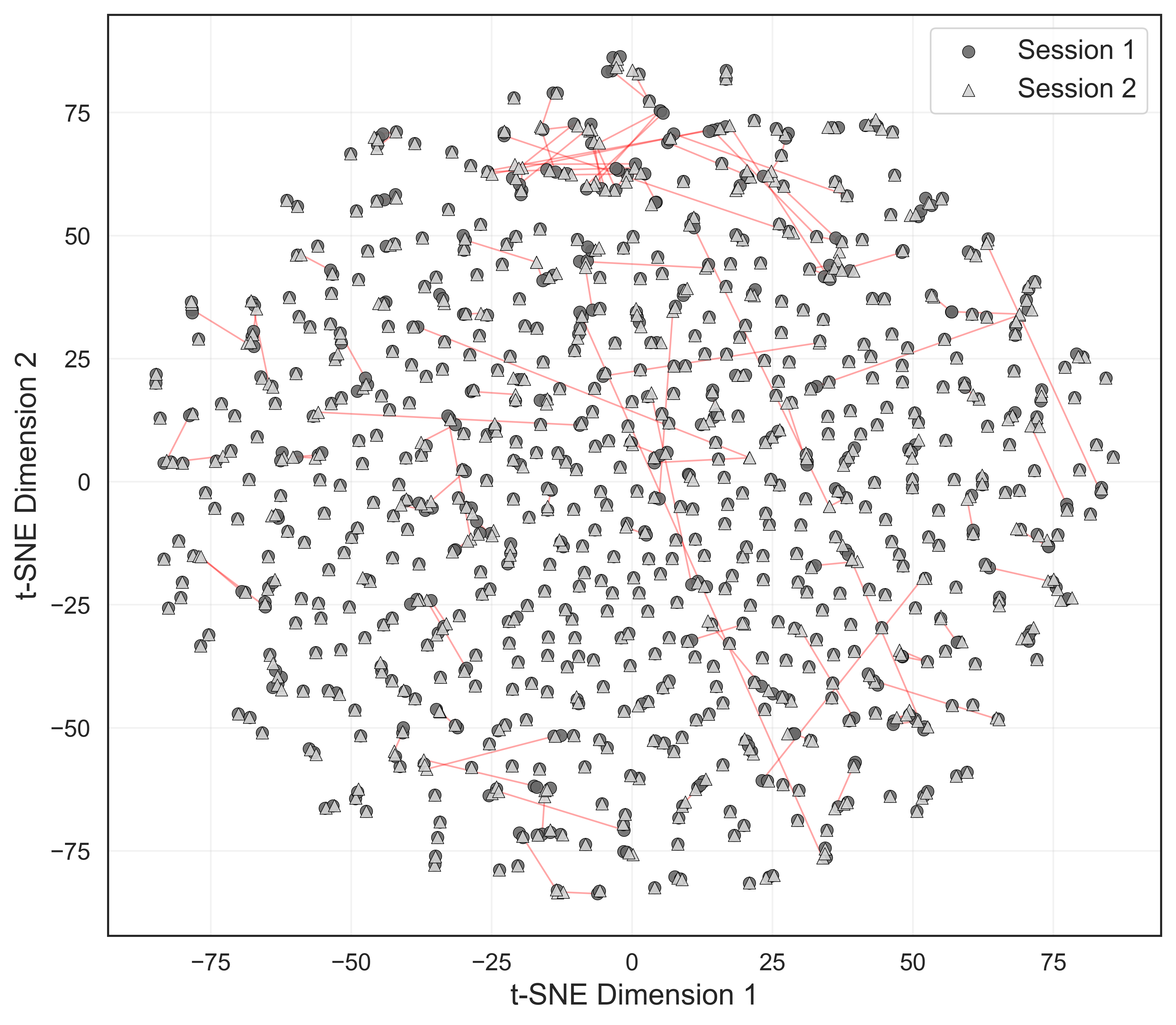}
    \caption{Best F2P}
    \label{fig:8b}
  \end{subfigure}
  \hfill
  \begin{subfigure}{0.24\linewidth}
    \includegraphics[width=\linewidth]{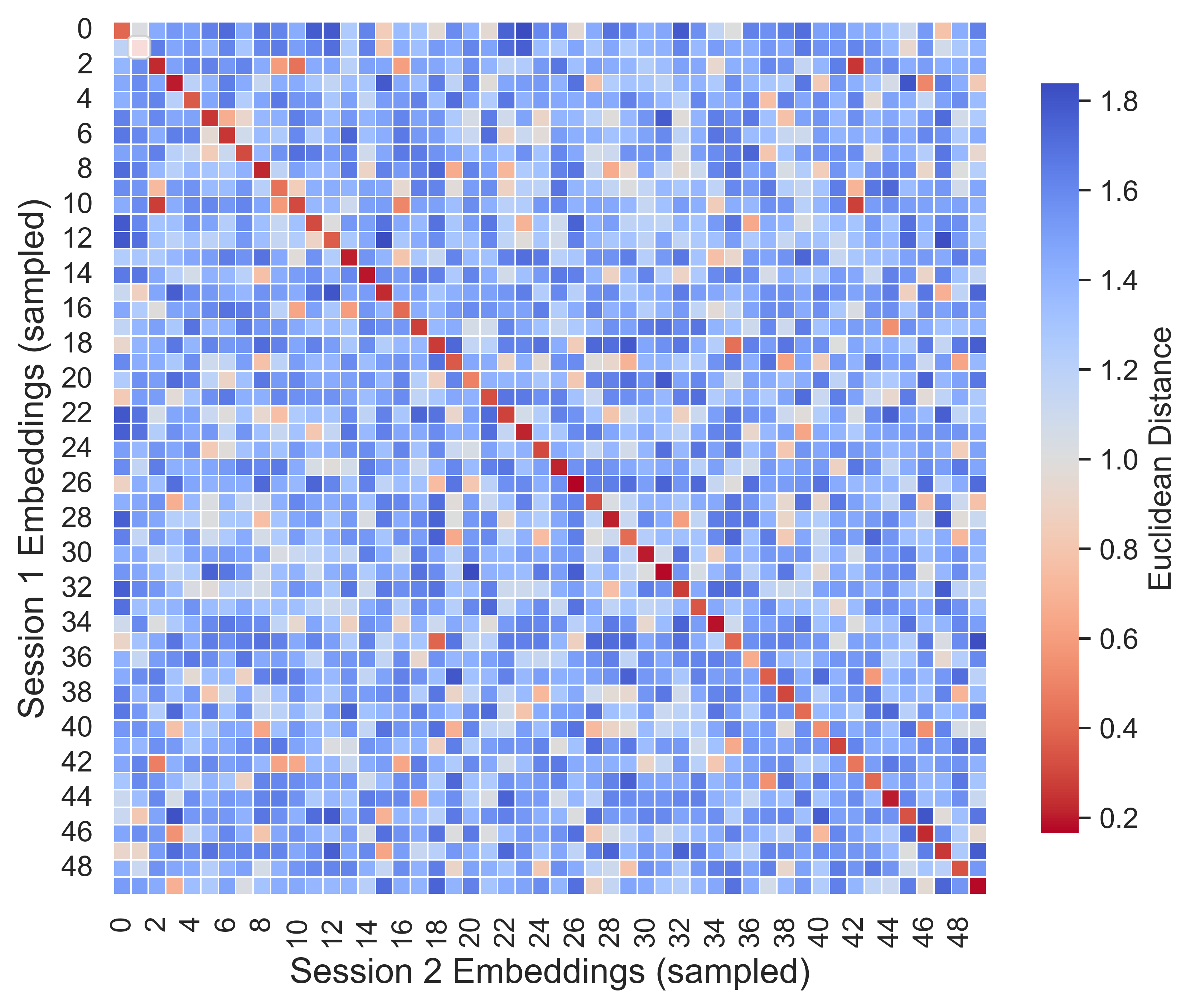}
    \caption{DeepPrint}
    \label{fig:8c}
  \end{subfigure}
  \hfill
  \begin{subfigure}{0.24\linewidth}
    \includegraphics[width=\linewidth]{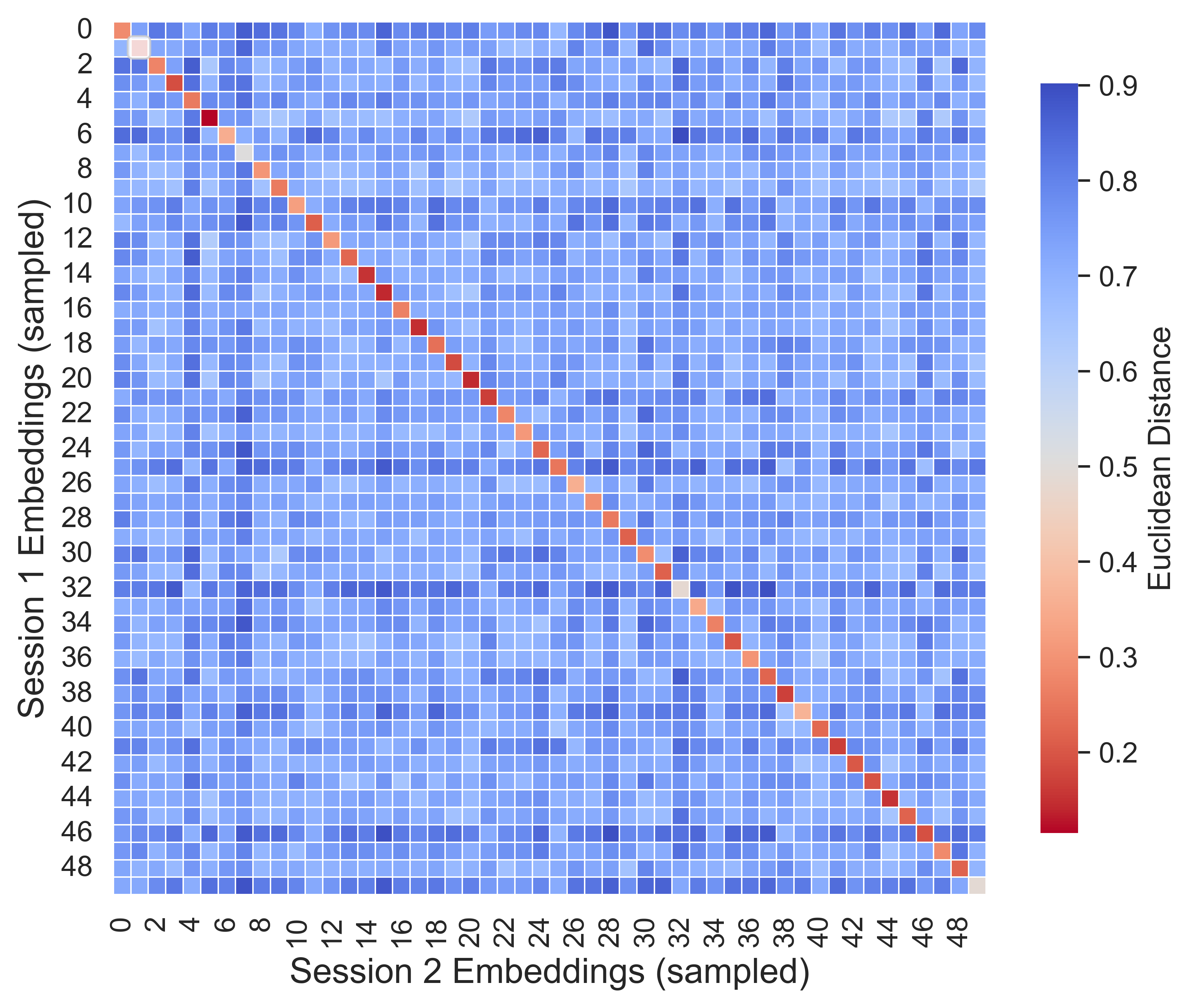}
    \caption{Best F2P}
    \label{fig:8d}
  \end{subfigure}

  \caption{\textbf{Comparison of DeepPrint Baseline and F2P representations.} (a),(b): t-SNE embedding space plots show DeepPrint's lower intra-subject consistency (more red links), while F2P achieves more stable features and higher verification accuracy. (c),(d): Pairwise embedding distance heatmaps show F2P with fewer off-diagonal matches, indicating stronger identity discrimination.}
  \label{fig:8}
\end{figure}

Fig.~\ref{fig:8} visualizes the embedding behavior of the DeepPrint baseline and the fine-tuned Fusion2Print with Spatial Transform variant.

\vspace{-2em}
\begin{table*}[!ht]
\centering
\caption{Comparison of DeepPrint Baseline and Fusion2Print variants with 128-D and 256-D embeddings, with and without spatial transform optimization. Metrics include intra-class distance (Intra), inter-class distance (Inter), and separation ratio (SepRatio = Inter/Intra), computed in both Euclidean and cosine similarity spaces. \textbf{F2P achieves a 1.66× improvement in SepRatio.}}
\label{tab:4}
\renewcommand{\arraystretch}{1.2}
\resizebox{\textwidth}{!}{%
\begin{tabular}{lccccccc}
\hline
\textbf{Model Variant} & \textbf{Embed Dim} &
\multicolumn{3}{c}{\textbf{Euclidean}} &
\multicolumn{3}{c}{\textbf{Cosine}} \\
\cline{3-5} \cline{6-8}
 &  & Intra$_{\downarrow}$ & Inter$_{\uparrow}$ & SepRatio$_{\uparrow}$ & Intra$_{\downarrow}$ & Inter$_{\uparrow}$ & SepRatio$_{\uparrow}$ \\
\hline
DeepPrint (Baseline Reference)  & 512 & 0.24 & 0.84 & 3.44 & 0.03 & 0.36 & 10.46 \\
DeepPrint w/ Spatial Transform  & 512 & \textbf{0.20} & 0.73 & 3.62 & \textbf{0.024} & 0.28 & 11.34 \\
F2P (Base)                      & 128 & 0.43 & 1.34 & 3.07 & 0.11 & 0.993 & 8.90 \\
F2P (Fine-tuned)                & 128 & 0.33 & 1.36 & 4.02 & 0.06 & 0.99 & 15.15 \\
F2P (Base)                      & 256 & 0.42 & 1.34 & 3.27 & 0.09 & 0.97 & 10.08 \\
F2P (Fine-tuned)                & 256 & 0.32 & 1.36 & 4.19 & 0.06 & \textbf{1.01} & 16.37 \\
\textbf{F2P (Fine-tuned) w/ Spatial Transform} & 256 & 0.31 & \textbf{1.37} & \textbf{4.30} & 0.05 & 0.995 & \textbf{17.38} \\
\hline
\end{tabular}
}
\end{table*}
\vspace{-2em}

\vspace{-2em}
\begin{table}
\centering
\caption{Verification performance for $I$ and $I_{\text{flash}}$ inputs across different embedding dimensions and F2P pipeline variants.}
\label{tab:5}
\setlength{\tabcolsep}{10pt}
\small
\resizebox{0.9\textwidth}{!}{%
\begin{tabular}{lccc}
\hline
\textbf{Embedding Dim} & \textbf{Pipeline Type} & \textbf{AUC} & \textbf{EER (\%)} \\
\hline
128 & Base      & 0.993 & 3.93 \\
    & Finetuned & 0.998 & 1.91 \\
\hline
256 & Base      & 0.997 & 1.97 \\
    & Finetuned & 0.999 & 1.30 \\
    & \textbf{Finetuned w/ Spatial Transform} & \textbf{0.999} & \textbf{1.12} \\
\hline
\end{tabular}%
}
\end{table}

\vspace{-2em}

\section{Conclusion}
\label{sec:conc}
We introduced \textbf{Fusion2Print (F2P)}, the first end-to-end contactless fingerprint verification framework that jointly performs flash--non-flash fusion, learned enhancement, and discriminative embedding generation. A key contribution is the \textbf{Flash--Non-Flash Fingerphoto (FNF) Database}, a custom paired dataset enabling systematic modeling of complementary flash and non-flash characteristics-previously underexplored in contactless fingerprint research.

Aligned flash--non-flash subtraction revealed ridge-preserving cues that improved minutiae localization. Building on this, \textit{DualEncoderFusionNet} adaptively combines illumination-robust features, and channel-wise analysis motivated the \textit{U-NetEnhancer} for channel-balanced ridge refinement. \textit{TripletDistilNet} produces compact 128-D/256-D embeddings that generalize across contact, contactless, and mixed-domain inputs, enabled by cross-domain fine-tuning.

End-to-end optimization of the enhancement--embedding pipeline yielded strong cross-session performance, with spatially normalized 256-D embeddings achieving the highest accuracy (\textbf{AUC 0.999}, \textbf{EER 1.12}), outperforming minutiae-based and single-capture baselines. 

F2P demonstrates that integrating flash--non-flash fusion, adaptive enhancement, and unified embedding learning enables accurate contactless and cross-domain fingerprint verification. A current limitation is the absence of adversarial attack robustness. Code has been released in the repository for reproducibility.

\subsection*{Future Work}
We plan to release a multi-device full-fingerprint dataset for broader cross-sensor evaluation. Future efforts could include exploring photometric stereo–based multi-illumination capture to enhance ridge visibility, and integrating the proposed approach with spoof detection systems for improved security and reliability.

%
%
%
\bibliographystyle{splncs04}
\bibliography{main}
%





\end{document}